\pdfoutput=1

\documentclass[11pt]{article}

\usepackage{arxiv}

\usepackage{times}
\usepackage{latexsym}

\usepackage[T1]{fontenc}

\usepackage[utf8]{inputenc}
\usepackage[square, comma, sort&compress, numbers]{natbib}
\usepackage{microtype}

\usepackage{inconsolata}

\usepackage{graphicx}

\usepackage{bm}
\usepackage{amsmath,amsfonts,amssymb}
\usepackage{amsmath,amssymb}
\usepackage{bm}
\usepackage{amsthm}
\usepackage{graphicx,color,epsfig,epstopdf}
\usepackage{algorithm}
\usepackage{algorithmic}
\usepackage{threeparttable}
\usepackage{multirow}
\usepackage{booktabs}   

\usepackage{hyperref}       
\usepackage{url}            
\usepackage{subfigure}
\usepackage{authblk}

\newcommand\numberthis{\addtocounter{equation}{1}\tag{\theequation}}

\newcounter{ass-counter}
\newcounter{thm-counter}
\newcounter{remark-counter}
\newtheorem{theorem}[thm-counter]{Theorem}

\newcommand{\alg}{{RazorAttention }}
\newcommand{\blg}{{RazorAttention}}

\title{RazorAttention: Efficient KV Cache Compression Through Retrieval Heads}
\author[1]{Hanlin Tang\thanks{Corresponding author: tanghl1994@gmail.com}}
\author[1]{ Yang Lin}
\author[1]{  Jing Lin}
\author[1]{ Qingsen Han}
\author[1]{ Shikuan Hong}
\author[1]{Yiwu Yao}
\author[1]{ Gongyi Wang}
\affil[1]{Huawei Technologies Co., Ltd}

\begin{document}

\maketitle

\begin{abstract}
The memory and computational demands of Key-Value (KV) cache present significant challenges for deploying long-context language models. Previous approaches attempt to mitigate this issue by selectively dropping tokens, which irreversibly erases critical information that might be needed for future queries. In this paper,  we propose a novel compression technique for KV cache that preserves all token information. Our investigation reveals that: i) Most attention heads primarily focus on the local context; ii) Only a few heads, denoted as retrieval heads, can essentially pay attention to all input tokens. These key observations motivate us to use separate caching strategy for attention heads. Therefore, we propose \blg, a training-free KV cache compression algorithm, which maintains a full cache for these crucial retrieval heads and discards the remote tokens in non-retrieval heads. Furthermore, we introduce a novel mechanism involving a ``compensation token'' to  further recover the information in the dropped tokens.  Extensive evaluations across a diverse set of large language models (LLMs) demonstrate that \alg achieves a reduction in KV cache size by over 70\% without noticeable impacts on performance. Additionally, \alg is compatible with FlashAttention, rendering it an efficient and plug-and-play solution that enhances LLM inference efficiency without overhead or retraining of the original model.
\end{abstract}

\section{Introduction}\label{sec:intro}
Long-context large language models (LLMs) have significantly advanced capabilities in natural language processing across diverse tasks. However, the growth of the Key-Value (KV) cache under increasing input length has become the major bottleneck for deployment. There are been plenty of previous work designed to alleviate this problem by compressing the KV cache size, including quantization ~\citep{pmlr-v202-sheng23a,zhao2024atom,lin2024qserve}, token-dropping ~\citep{zhang2023ho,xiao2024efficient}, local attention ~\citep{mistral, sparseattn}, etc.


\begin{figure}[htbp]
\centering  
\includegraphics[width=0.49\textwidth]{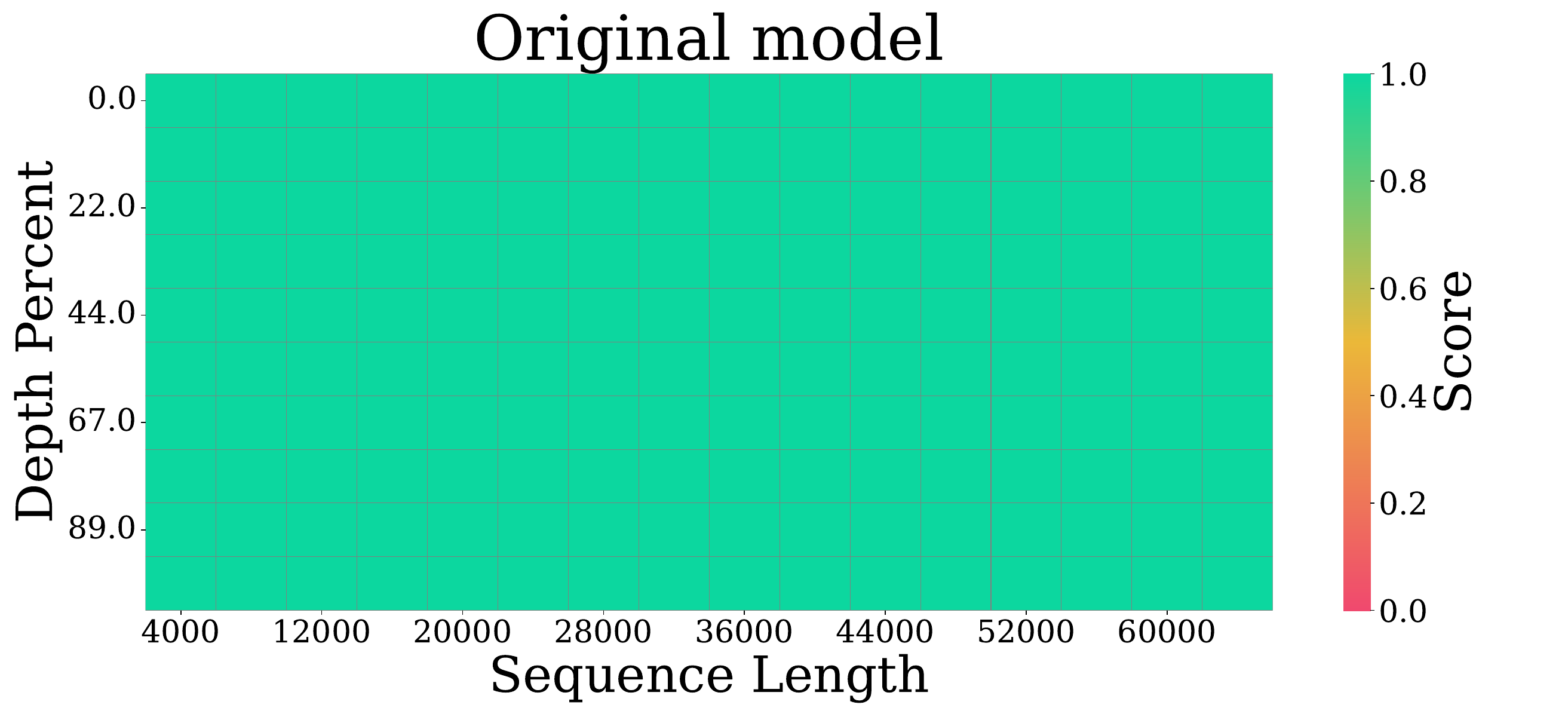}
\includegraphics[width=0.49\textwidth]{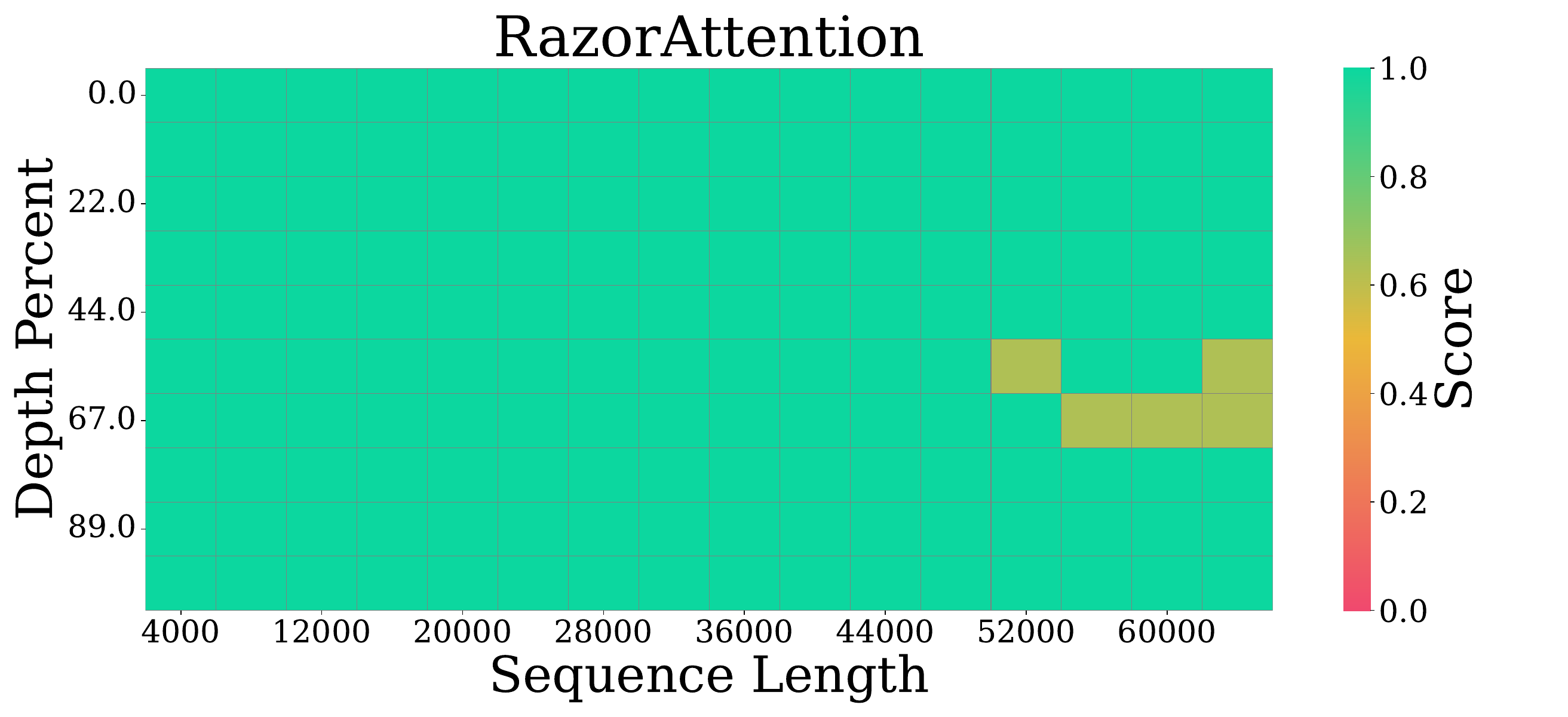}
\caption{\alg achieves comparable performance to the original model, even  with  $70\%$ KV cache compressed. To demonstrate this, we tested Llama2-13B-64K~\citep{fu2024data} on the Needle in A Haystack benchmark~\citep{gkamradt2023needle}.}
\label{fig:intro_needle}
\end{figure}

\begin{figure*}[t]
\centering 
\includegraphics[width=1.0\textwidth]{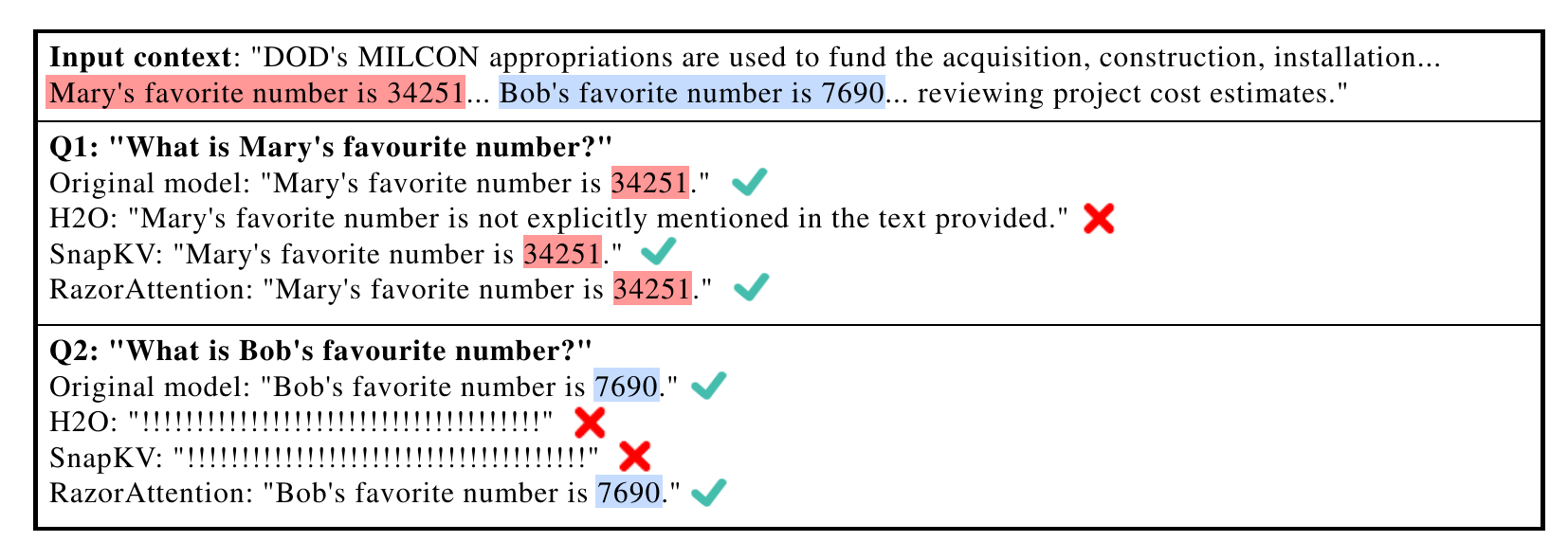} 
\caption{Importance-based token-dropping methods cannot work when querying the less relevant information to the main theme. Here, we use an 8K document from LongBench~\citep{bai2023longbench} and add two sentences that are not relevant to the main theme. In this case, H2O discards tokens that are less relevant to the main theme, leading to failures in both Q1 and Q2. SnapKV discards tokens based on the first query, making it effective for Q1 but failing in subsequent queries like Q2. Only \alg successfully outputs the exact information from the lengthy input even when we compress $70\%$ of the KV cache.}
\label{Fig.demo_conversation}
\end{figure*}

One major direction for KV cache compression is to directly drop tokens deemed unimportant so far~\citep{zhang2023ho,xiao2024efficient,liu2023scissorhands,li2024snapkv}.
These methods inherently assume that tokens considered unimportant will not be needed in future queries, which does not hold in practical scenarios. For instance, a user might request information that is not directly aligned with the main theme of the processed text, or engage in a multi-round conversation querying different segments from the context. In these cases,  the importance-based token-dropping methods can lead to significant performance degradation since the actual information required by the query  might be discarded if considered unimportant (see our example on Qwen1.5-7B-Chat~\citep{bai2023qwen} in Figure~\ref{Fig.demo_conversation}). This leads us to pose a critical question:
\begin{center}
\textit{``Can we find a way to reduce the KV cache size without losing semantic information?''}
\end{center}

In this work we address this problem from a novel perspective.  Our investigation reveals that there  exists a ``retrieve and process'' mechanism in LLMs when processing a long context. More specifically, LLMs can accurately recall the queried information from a lengthy input through certain group of attention heads, which we denote as ``retrieval heads'' (see Section~\ref{sec:def_retrieval} for definition). These heads are capable of concentrating most of their attention weights on the relevant information (w.r.t. the queries) and increasing the output probability for those words. 
Another important finding is that non-retrieval heads primarily focus on local context or the attention sink~\citep{xiao2024efficient}, which means these heads cannot effectively utilize all the semantic information from the input.
Based on these important findings, we hypothesize that LLM runs the reasoning procedure on a ``retrieve and process'' basis. That says, the model first uses the retrieval heads to gather relevant information, and then non-retrieval heads to process the retrieved information and generate the final response. This motivates us to design separate caching strategies for different heads: For retrieval heads, we keep the KV cache unaltered; for the rest heads, we only cache  recent tokens and attention sinks. 

Beyond this, we notice that there still exists a certain accuracy gap when we directly discard all the remote tokens in the non-retrieval heads. Therefore for these non-retrieval heads, we designed a ``compensation token'' for compressing the dropped cache into one token, and proved that the accuracy degradation due to the truncated KV cache gets further improved with this compensation token. With retrieval heads and compensation tokens, we prove that our algorithm, namely \blg, can successfully compress $70\%$ of the KV cache without noticeable performance degradation as illustrated in Figure~\ref{fig:intro_needle}.


Last but not least, previous importance-based token-dropping methods  cannot be combined with FlashAttention due to their reliance on the attention weights to compute the importance score, making them impracticable for implementation since FlashAttention is one of the most important components in long-context inference. \alg addresses this  problem since it does not use the attention map as  the metric. The head-wise pruning criterion is totally compatible with FlashAttention, and the computation overhead of the compensation token is negligible. Therefore \alg could achieve a substantial inference speedup   when compared to previous methods. 

To the best of our knowledge, \alg is the first training-free token reduction algorithm that achieves a nearly lossless 3X KV cache reduction. We evaluated \alg on models including Qwen~\citep{bai2023qwen}, Llama-2~\citep{llama2}, Llama-3~\citep{llama3modelcard} and Baichuan~\citep{baichuan2023baichuan2} on long-context tasks to prove its effectiveness. Our contribution can be summarized as follows:
\begin{itemize}
\item We systematically analyze the attention dynamic of Transformers under lengthy inputs. Our work reveals that only a few retrieval heads can essentially recall information from the whole input while the rest heads mainly focus on the local context. 
\item We introduce a novel algorithm, namely \blg, that is capable of reducing the KV cache size by $70\%$ under minimal impact on performance for contexts ranging from 8K to 100K tokens. We designed an accurate and data-free metric for allocating all the retrieval heads, together with an error compensation strategy for compensating the information loss due to the truncated KV cache.
\item  \alg introduces negligible overhead in compression and is compatible with FlashAttention,  rendering it an efficient and plug-and-play solution that enhances LLM inference efficiency without training or significant overhead. Extensive experiments demonstrate that \alg can be effectively applied to various models and tasks.
\end{itemize}

\section{Related Work}
As the sequence length increases, the memory consumption of  KV cache rapidly expands, potentially surpassing the size of the model parameters themselves. This leads to an urgent need for KV cache compression, particularly in scenarios with limited GPU memory. One direction  is non-Transformer architecture design, such as Mamba~\citep{mamba}, Mamba2~\citep{mamba2}, Infini-Transformer~\citep{infinitransformers}, RWKV~\citep{rwkv}and Griffin~\citep{griffin}. However, in this paper we focus on KV cache reduction for typical Transformers, which is the most widely used model structure. Below we introduce several approaches for KV cache compression.

\paragraph{Quantization} Quantization is a classic yet effective approach to neural network compression. In the field of LLM Quantization, while the outlier challenge attracts great attention~\citep{pmlr-v202-xiao23c, NEURIPS2022_6f6db140, wei2023outlier} to tackle, the application of which on KV cache is often seen as a by-product of activation quantization. Nevertheless,  there are several noteworthy works demonstrating the value of KV cache quantization.  FlexGen, Atom and QServe~\citep{pmlr-v202-sheng23a,zhao2024atom,lin2024qserve} carefully designed quantization pipelines that utilize KV cache compression to boost the overall inference throughput. KVQuant~\citep{hooper2024kvquant} integrates several techniques to minimize KV quantization error and KIVI~\citep{liu2024kivi} pushed the limit towards 2-bits. Besides the post-training methods, LLM-QAT~\citep{liu2023llm} offers a data-free distillation process that further recovers the performance of the model.
 
\paragraph{Token-dropping}  Token-dropping methods assume that not all key-value pairs are essential in self-attention computations, so memory usage can be saved by identifying and removing unimportant KV cache. StreamingLLM~\citep{xiao2024efficient} utilizes sliding window technology, preserving only the KV pairs of attention sink tokens and those within the sliding window, thereby reducing memory footprint and stabilizing model performance. H2O~\citep{zhang2023ho}
is one of the pioneers that use the attention scores to evaluate the importance of each token, followed by an eviction strategy that greedily selects cache with higher scores.
Scissorhands~\citep{liu2023scissorhands} and one of the latest work SnapKV~\citep{li2024snapkv} use similar ideas by narrowing the computation range to consider attention scores related to recent information. Built on that, PyramidKV and PyramidInfer~\citep{cai2024pyramidkv,yang2024pyramidinfer} analyze the attention concentration patterns and further reduce KV cache in later layers. Moreover, research efforts have been made to understand KV cache from different perspectives: FastGen~\citep{ge2024model} paid attention to special tokens and punctuation, SubGen~\citep{zandieh2024subgen} investigated the clusterability of key embedding and CORM~\citep{dai2024sequence} discovered strong correlation amongst tokens of near neighbors. 


\paragraph{Non-MHA Attention} Another category focuses  on reducing KV cache by sharing cache across attention heads. MQA~\citep{shazeer2019fast} aggressively uses a single KV head for all heads, whereas GQA~\citep{ainslie2023gqa} suggests an intermediate number of heads to balance the trade-off between inference speed and output quality. Furthermore, MLA~\citep{deepseekai2024deepseekv2} presents a novel caching method by low-ranking KV cache of all heads into single latent space.

Our algorithm is motivated by the idea from ~\cite{olsson2022incontext}, where the authors noticed that there are certain groups of attention heads, denoted as the induction heads, that can effectively recall the queried information from the input. Recent study~\citep{wu2024retrieval} also validated this property under extended inputs. This is the first work that proposes  a head-wise pruning criterion for KV cache compression based on the interpretability of the attention mechanism.

\section{Methodology}
In this section, we introduce the key components of \blg. We firstly apply \alg to models using ALiBi~\citep{alibi} positional embedding (denoted as ALiBi models) to provide an intuitive understanding of the retrieval and non-retrieval heads. Afterwards, we demonstrate that models using RoPE~\citep{rope} positional embedding (denoted as RoPE models) also exhibit this crucial characteristic, which reveal that KV cache within RoPE models can also be efficiently compressed under minimal loss of accuracy. 

\subsection{\alg for ALiBi models}
For ALiBi models, its $h$-th  attention head computes the attention score according to 
\begin{align*}
S_{m\to n}\left(\bm{q};\bm{k}  \right) = \bm{q}_m\bm{k}_n^{\intercal} - l_h (m-n),\numberthis\label{eq:ALiBi_score}
\end{align*}
where  $\bm{q}_m$ is the query tensor at the $m$-th position, $\bm k_n$ is the key tensor at the $n$-th position, $l_h$ is the head-specific slope, $S_{m\to n}\left(\bm{q};\bm{k}  \right)$ is the attention score. Notice that $(m\geq n)$ is guaranteed by the casualty of attention. 

In the scenario where $l_h(m-n)$ significantly dominates $\bm{q}_m\bm{k}_n^{\intercal}$, the attention between $\bm{q}_m$ and $\bm{k}_n$ would decay to zero, meaning that the contribution of any tokens positioned further than $n$ becomes negligible for the output at position $m$. The following theorem formalizes this observation.
\begin{theorem}\label{theo:ALiBi_theo}
Given an attention head that calculates the attention score as per \eqref{eq:ALiBi_score}, for any $\epsilon\in (0,1)$, the attention weight from $\bm{q}_m$ to $\bm{k}_n$ can be upper bounded by:
\begin{align*}
&\text{Attn}_{m\to n}\left(\bm{q};\bm{k}  \right)
= \frac{\exp{\left(S_{m\to n}\left(\bm{q};\bm{k}  \right) \right)}}{\sum_{n=0}^{m}\exp{\left(S_{m\to n}\left(\bm{q};\bm{k}  \right) \right)}} \leq \epsilon,\quad\forall n < m - C_0,\\
&L_h:=\frac{2\|W_{Q_h}W_{K_h}\|_2\left( \|\bm \gamma\|^2 + \|\bm b\|^2 \right) - \log(\epsilon)}{l_h}.\numberthis\label{eq:ALiBi_theo}
\end{align*}
Here $W_{Q_h}$ and $W_{K_h}$ are the query and key matrices of the $h$-th attention head, $\gamma$ and $\bm{b}$ are the weight and bias for the LayerNorm layer before attention ($\bm{b}=\bm{0}$ for RMSNorm~\citep{rmsnorm}), and $\|\cdot\|_2$ denotes the $l_2$-norm of the matrix. $L_h$ can be viewed as the vision scope of the head. The detailed proof can be found in Appendix~\ref{sec:proof}.
\end{theorem}
Theorem~\eqref{theo:ALiBi_theo} indicates that when the distance between $\bm{q}_m$ and $\bm{k}_n$ exceeds $C_0$, the attention weight between these two tokens falls below $\epsilon$. When $\epsilon$ is sufficiently small (e.g., 0.1\%), remote tokens impose minimal influence on the final output and can thus be discarded. Building on this principle, ALiBi models dynamically adjust the KV cache size for each head. We first compute the effective attention scope $L_h$, and keep only the recent $L_h$ tokens in the KV cache, since any token further than $L_h$ impose attention weight no more than $\epsilon$, we can safely discard them for compression. Therefore, for ALiBi models, the retrieval heads are the ones with a larger $L_h$, while the non-retrieval heads has a  smaller attention vision $L_h$.

\subsection{\alg for RoPE models}
For RoPE models, each attention head computes the attention score according to 
\begin{align*}
&S_{m\to n}\left(\bm{q};\bm{k}  \right) = \bm q_m \bm k_n^{\intercal},\quad\bm q_m=\mathcal{R}_m\bm q, \quad  \bm k_n=\mathcal{R}_n\bm k
 \numberthis\label{eq:def_rope}
\end{align*}
where $\bm q_m$ and $\bm k_n$ are the query and key state after rotary transformation,  $\mathcal{R}_m$ and $\mathcal{R}_n$ are the rotary matrices at position $m$ and $n$ (see \cite{rope} for details). Although RoPE embedding does not inherently suggest a long-range decaying attention, our empirical findings indicate that the majority of attention heads maintain a limited scope of attention. Notably, only about 15\% of the heads, which we term as retrieval heads, are capable of effectively utilizing long-range information while the rest heads only focus on the local context. As shown in Table~\ref{tab:protect}, a significant decrease in accuracy of 16\% is observed when the KV cache size is reduced for these retrieval heads. Conversely, dropping remote tokens within non-retrieval heads results in a comparatively minor performance degradation of 1.5\%. 
\begin{table}
\centering
\begin{tabular}{|c|c|c|c|c|}
\hline
Protection heads & All & Retrieval heads &Random heads & None  \\\hline
MultiFieldQA-en & 46.94\% & 45.48\% & 40.7\% & 40.81\%\\\hline
\end{tabular}
\caption{\label{tab:protect}We protected the KV cache within different groups of attention heads while keeping only the recent 4K tokens in the rest. 
Protecting the KV cache in the retrieval heads  can retain most of the LLM's performance, while protecting randomly heads brings no performance gain. This clearly indicates that most of the attention heads only use local context and only retrieval heads can essentially utilize all context information.}
\end{table}

\begin{figure}[htbp]
        \centering
        \subfigure[]{\label{fig:show_rope_ra}
      \includegraphics[width=0.48\textwidth]{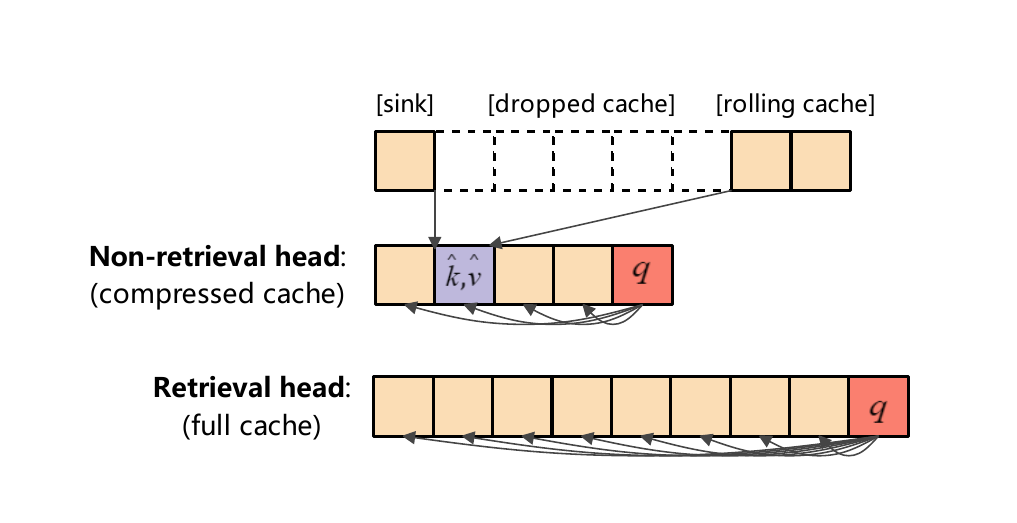}
        }
        \quad
        \subfigure[]{\label{fig:induction+echo}
        \includegraphics[width=0.4\textwidth]{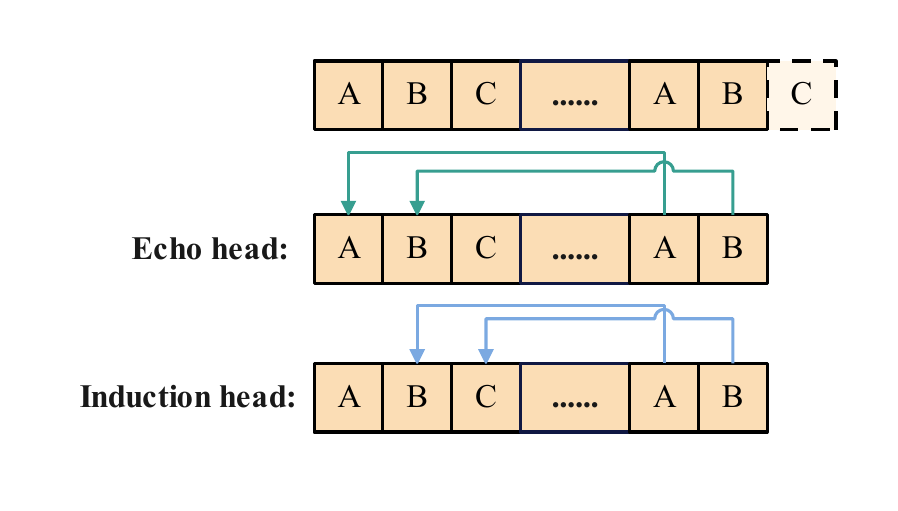}}
        \caption{In Figure~\ref{fig:show_rope_ra} we present the illustration of how \alg compress the KV cache. For retrieval heads, we maintain a full cache for retaining all the tokens' information. For non-retrieval heads, we directly discard remote tokens and compress the discarded tokens into a compensation token whose KV cache is denoted as $\{\hat{\bm{k}}, \hat{\bm{v}}\}$. In Figure~\ref{fig:induction+echo} we provide an illustration  example of the echo head and induction head. The current token is ``B'' and the generated token is ``C''. In this case, the echo head would mainly attend to token ``B'' while the induction head mainly attend to token ``C'' in previous context.}\label{fig:ra_illustration}
\end{figure}

Based on the findings above, we directly decrease the KV cache for all non-retrieval heads. The performance of the model is mostly retained as shown in Table~\ref{tab:protect}.However, a notable accuracy gap remains, indicating that some information is still being lost. Moreover, the test result on Needle in a Haystack shows a clear performance degradation even when we protect the KV cache of retrieval heads (see our ablation result in Figure~\ref{fig:ablation_compenation_token}). To further improve performance, we designed a lightweight and effective way to compress the information in the dropped token into a ``compensation token''. The compensation token is  defined as
\begin{align*}
\hat{\bm{k}} = \frac{1}{N_{d}}\sum_{m\in \{\mathcal{D}\}} \bm{k}_m,\quad \hat{\bm{v}} = \frac{1}{N_{d}}\sum_{m\in \{\mathcal{D}\}} \bm{v}_m.\numberthis\label{eq:compensation token}
\end{align*}
Here $\hat{\bm{k}}$, $\hat{\bm{v}}$ are the compensation tokens for the dropped KV cache, $\{\mathcal{D}\}$ contains the indices of the dropped tokens and  $N_{d}$ is the number of the dropped tokens. Afterward, we discard the dropped tokens and augment the KV cache with the compensation token $\hat{\bm k}$ and $\hat{\bm v}$, where $\{K, V\}$ are the KV cache of the remaining token after rotary transformation. Denoting the compressed KV cache as $\{K,\hat{\bm{k}}\}$ and $\{V,\hat{\bm{v}}\}$, the attention output of the current token follows
\begin{align*}
&\text{Attn}(\bm{q}_m, \{K,\hat{\bm{k}}\}, \{V,\hat{\bm{v}}\}\})
=\frac{N_{d}\exp{\left( \bm{q}_m\hat{\bm{k}}^{\intercal}\right)}\hat{\bm v} + \sum_{n\notin \{\mathcal{D}\} }\exp{\left( \bm{q}_m\bm {\bm{k}}_n^{\intercal}\right)}{\bm{v}}_n}{N_{d}\exp{\left(\bm{q}_m\hat{\bm{k}}^{\intercal}\right)} + \sum_{n\notin \{\mathcal{D}\} }\exp{\left( \bm{q}_m\bm {\bm{k}}_n^{\intercal}\right)}}.\numberthis\label{eq:ra_update_attention}
\end{align*}
In Figure~\ref{fig:show_rope_ra} we provide an illustrative example of \alg for RoPE models. With compensation tokens, the accuracy is further improved, making \alg almost lossless even dropping $70\%$ of the KV cache in the non-retrieval heads. Below we introduce how we determine the retrieval heads group.


\subsection{Identification of Retrieval Heads}\label{sec:def_retrieval}
For ALiBi models, the attention scope can be directly determined via \eqref{eq:ALiBi_theo} and KV cache can be dropped accordingly. However, for RoPE models, the retrieval heads need to be identified in a more sophisticated way.
Our investigation reveals that two groups of heads are essential in processing long context, so both of them should be included as retrieval heads as stated below.
\begin{itemize}
\item \textbf{Echo head}: The head tends to attends back to previous token (referred as echo token) identical to the current token.
\item \textbf{Induction head}: The head tends to attend to the previous token (namely induction token) that is immediately succeeded by the current token. Basically it attends to the coming token that also exists in previous context.
\end{itemize}
In Figure~\ref{fig:induction+echo} we present an illustrative example explaining the echo heads and induction heads. In order to identify the retrieval heads, we generate $K$ (for example, $K=2500$) random tokens, repeat these tokens 4 times, and then use it as the input of the model. This design minimizes semantic dependencies among tokens, thereby allowing a clearer observation of the behavior of echo and induction heads.

Subsequently, we calculated the echo score (attention weight to the echo token) and induction score (attention weight to the induction token) of all words across all heads. The selection of retrieval heads involves the top-$14\%$ attention heads with the highest induction score and top-$1\%$ of attention heads with the highest echo score (see Table~\ref{tab:hyperparameters}). Notice that although we only use much fewer echo heads than retrieval heads, our investigation indicates that both heads are crucial for the retrieving performance for LLMs (see Section~\ref{sec:ablation} for ablation results).

With the retrieval heads being identified, we hereby introduce \alg for RoPE Models in Algorithm~\ref{alg:rope}.
\begin{algorithm}[H]
\caption{\alg for RoPE Models}
\renewcommand{\algorithmicrequire}{\textbf{Input:}}
\renewcommand{\algorithmicensure}{\textbf{Output:}}
\label{alg:rope}
\begin{algorithmic}[1]
 \REQUIRE Non-retrieval headset $\{H\}$, original KV cache (after rotary transformation) $\{K, V\}$, compression ratio $C$, compression threshold $S_0$, sink token num $N_0$.\\
\FOR{non-retrieval head $h\in \{H\}$}
\STATE Compute the buffer length $L_h = \max\left(S_0, \frac{N}{C}\right)$, here $N$ is the number of tokens in the head.\;
\STATE Keeping only the recent $L_h$ tokens near output and first $N_0$ sink tokens, discarding the remaining tokens and compress them into a compensation token according to \eqref{eq:compensation token}.\;
\ENDFOR
\STATE Non-retrieval heads compute attention according to \eqref{eq:ra_update_attention}, while retrieval heads follow the original attention.
\ENSURE  Generated output tokens.
\end{algorithmic}  
\end{algorithm}

\begin{table}
\centering
\begin{tabular}{|c|c|}
\hline
Hyper-parameter & Settings \\\hline
Buffer length & $\max(4000, N/5)$\\
Induction head protection & top $14\%$\\
Echo head protection & top $1\%$\\
Sink token num & 4 \\
\hline
\end{tabular}
\caption{\label{tab:hyperparameters}General hyper-parameter settings for experiments in the paper, which leads to 3.125x compression of KV cache under long context input.}
\end{table}

\begin{table*}[t]

\fontsize{21}{28}\selectfont
\setlength{\tabcolsep}{5pt}
\centering

\begin{threeparttable}

\scalebox{0.3}{
\begin{tabular}{l|lcccccccccccccccc}
\specialrule{1pt}{0pt}{2pt}
LLMs&& \rotatebox[origin=c]{30}{NrtvQA} & \rotatebox[origin=c]{30}{Qasper} & \rotatebox[origin=c]{30}{MF-en} & \rotatebox[origin=c]{30}{MF-zh} & \rotatebox[origin=c]{30}{HotpotQA} & \rotatebox[origin=c]{30}{2WikiMQA} & \rotatebox[origin=c]{30}{Musique} & \rotatebox[origin=c]{30}{GovReport} & \rotatebox[origin=c]{30}{QMSum} & \rotatebox[origin=c]{30}{MultiNews} & \rotatebox[origin=c]{30}{VCSUM} & \rotatebox[origin=c]{30}{TREC} & \rotatebox[origin=c]{30}{TriviaQA} & \rotatebox[origin=c]{30}{LSHT} & \rotatebox[origin=c]{30}{Lcc}& \rotatebox[origin=c]{30}{Average}\\

\specialrule{1pt}{2pt}{2pt}

\multirow{4}{*}{\rotatebox[origin=c]{90}{\fontsize{22}{100}\selectfont Qwen1.5-}\rotatebox[origin=c]{90}{\fontsize{22}{100}\selectfont 7B-Chat}}

&~~~All KV & 17.58 & 43.16 &  46.94 & 60.98 & 50.96 & 36.36& 27.86 & 28.78 & 23.24 & 24.02 & 13.91 &17.64& 83.77 & 16.96 & 48.35 & 36.03\\
\cline{2-18}

&~~~StreamingLLM & 6.22&24.62& 18.9 & 34.51 & 20.68 & 12.31& 5.88 & 3.86 & 3.52 & 20.74 & 3.17 & 8.5 & 36.57 & 13 & 42.51 & 17.00\\

&~~~H2O & 16.5&38.15& 40.22 & 51.46 & 50.19 & 35.69& 27.12 &\textbf{ 28.42} & 22.00 & 22.70 & \textbf{14.03} & 18.25 & 83.72 & \textbf{16.4} & 47.54 & 34.16\\

&~~~RA & \textbf{16.63}&\textbf{43.1}& \textbf{46.66} & \textbf{61.08} & \textbf{50.49} & \textbf{36.1}& \textbf{28.79} & 26.68 & \textbf{22.59} & \textbf{23.96} & 13.83 & \textbf{20.87} & \textbf{83.83} & 15.66 & \textbf{47.85 } & \textbf{35.87}\\

\specialrule{1pt}{2pt}{10pt}\specialrule{1pt}{2pt}{2pt}

\multirow{4}{*}{\rotatebox[origin=c]{90}{\fontsize{20}{22}\selectfont Qwen1.5-}\rotatebox[origin=c]{90}{\fontsize{20}{22}\selectfont 72B-Chat}}

&~~~All KV & 28.32 & 46.73 & 48.25 & 63.41 & 55.91 & 46.23 & 34.56 & 32.47 & 22.69 & 24.86 & 15.61 & 71.0 & 91.15 & 46 & 65.05 & 46.15\\
\cline{2-18}
&~~~StreamingLLM & 9.57 & 28.33 & 19.06 & 34.98 & 25.32 & 13.42 & 10.08 & 4.11 & 3.79 & 21.1 & 3.74 & 43.0 & 43.72 & 20.5 & 53.6 & 22.29\\

&~~~H2O &\textbf{ 27.98} & 41.45 & 43.69 & 55.93 & 54.77 & 45.16 & \textbf{34.61} & 32.24 & 22.35 & 24.36 & 14.5 & 70.0 & 91.15 & 42 & 64.2 & 44.29\\

&~~~RA & 27.97 & \textbf{46.44} & \textbf{47.36} & \textbf{63.04} & \textbf{55.92 }& \textbf{46.15} & 34.36 &\textbf{ 32.35 }& \textbf{22.75} & \textbf{24.91} & \textbf{15.17} & \textbf{71.0} & \textbf{91.49} & \textbf{46} & \textbf{64.68} & \textbf{45.97}\\

\specialrule{1pt}{2pt}{10pt}\specialrule{1pt}{2pt}{2pt}

\multirow{4}{*}{\rotatebox[origin=c]{90}{\fontsize{22}{100}\selectfont Llama3-8B}\rotatebox[origin=c]{90}{\fontsize{22}{100}\selectfont -Instruct}}

&~~~All KV & 21.84 & 37.04 & 45.07 & 52.34 & 44.63 & 27.28 & 23.04 & 28.18 & 24.54 & 26.26 & 14.41 & 0 & 85.90 & 3 & 30.17 & 35.44\\
\cline{2-18}
&~~~StreamingLLM & 0.61 & 16.29 &  13.41 & 20.05 & 2 & 5.84& 0.37 & 5.22 & 4.63 & 18.89 & 2.52 &-& 11.54 & - & 26.83 & 9.86\\

&~~~H2O & 21.14 & 34.1 & 40.84 & 47.13 & 43.47 & \textbf{27.13} & 21.31 & 22.85 & 16.36 & 22.3 & 14.52 & - & \textbf{86.17} & - & \textbf{30.26} & 32.89\\

&~~~RA & \textbf{21.16} & \textbf{36.22} & \textbf{42.88} & \textbf{51.93} & \textbf{44.07} & 26.89 & \textbf{22.03} & \textbf{26.56} & \textbf{23.86} & \textbf{25.83} & \textbf{15.69} & - & 85.83 & - & 30.25 & \textbf{34.86}\\

\specialrule{1pt}{2pt}{10pt}\specialrule{1pt}{2pt}{2pt}

\multirow{4}{*}{\rotatebox[origin=c]{90}{\fontsize{22}{100}\selectfont Baichuan2}\rotatebox[origin=c]{90}{\fontsize{22}{100}\selectfont -13B}}

&~~~All KV & 18.63 & 30.16 & 44.1 & 50.36 & 37.93 & 32.62& 13.90 & 26.16 & 20.14 & 24.58 & 15.66 & 62.5 & 86.61 & 27.5 & 55.36 & 36.41\\
\cline{2-18}
&~~~StreamingLLM & 5.12 & 12.44 &  23.53 & 32.52 & 16.93 & 16.08 & 6.15& 5.53 & 1.03 & 5.6 & 3.94 &42.22& 30.15 & 7.32 & 35.42 & 16.27\\

&~~~H2O & 17.81 & 29.89 & \textbf{43.74} & 49.54 & \textbf{37.02} & 31.71& 13.54 & \textbf{25.8} & 18.96 & 23.31 & 15.11 & 62.41 & 85.25 & 26.86 & 54.45 & 35.69\\

&~~~RA & \textbf{18.22} & \textbf{31.87} & 43.6 & \textbf{51.36} & 36.97 &\textbf{ 32.89}& \textbf{13.98} & 25.51 & \textbf{20.13} & \textbf{24.51} & \textbf{15.41} & \textbf{62.5} & \textbf{87.23} & \textbf{28} & \textbf{54.53} & \textbf{36.45}\\

\specialrule{1pt}{2pt}{10pt}

\end{tabular}

}
\end{threeparttable}
\caption{Performance comparison of \alg and other compression algorithms across various LLMs on LongBench. Notice that the performance of Llama3-8B-Instruct on TREC and LSHT are not applicable (close to 0), hence we do not include their result on Llama3-8B.}\label{tab:longbench}

\end{table*}

\begin{figure*}[t]
\centering  
\includegraphics[width=0.32\textwidth]{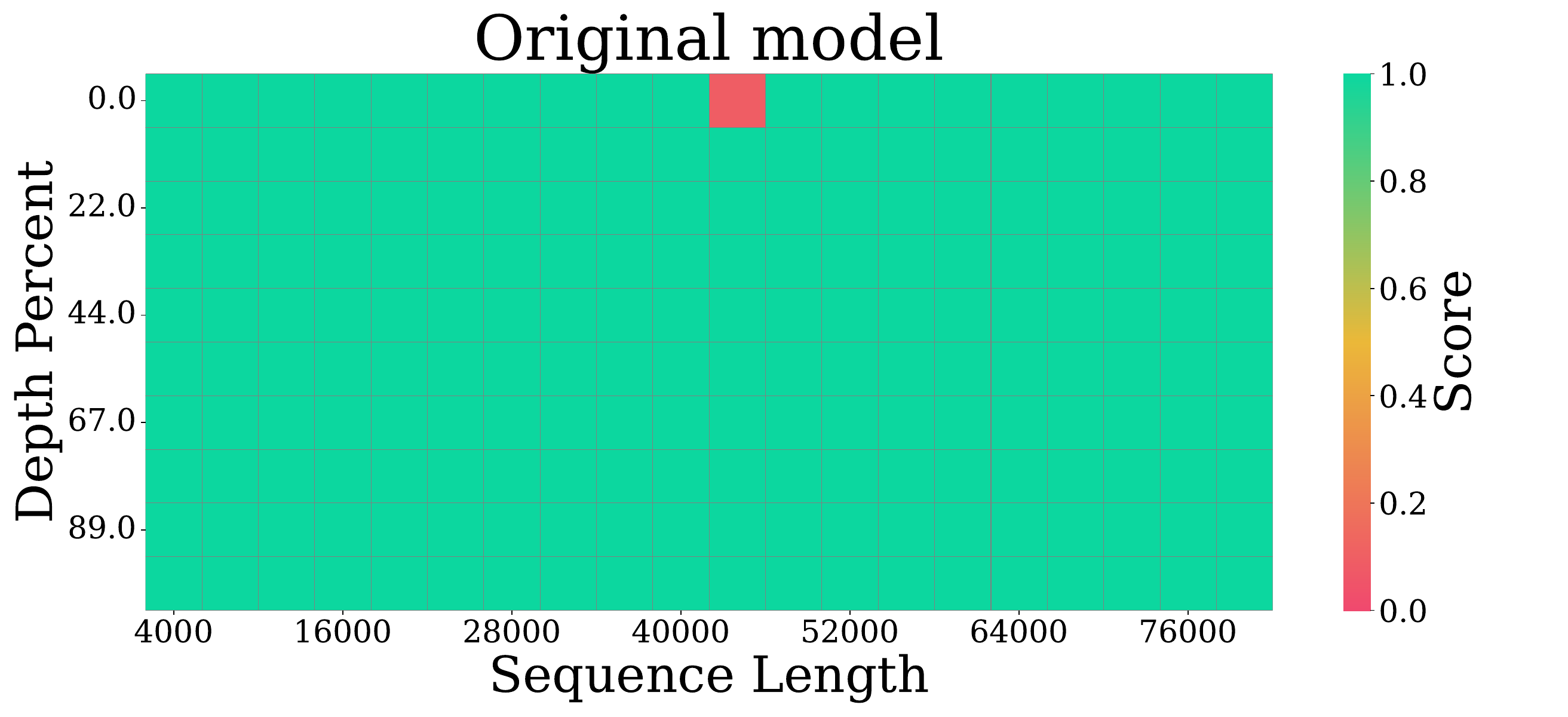}
\includegraphics[width=0.32\textwidth]{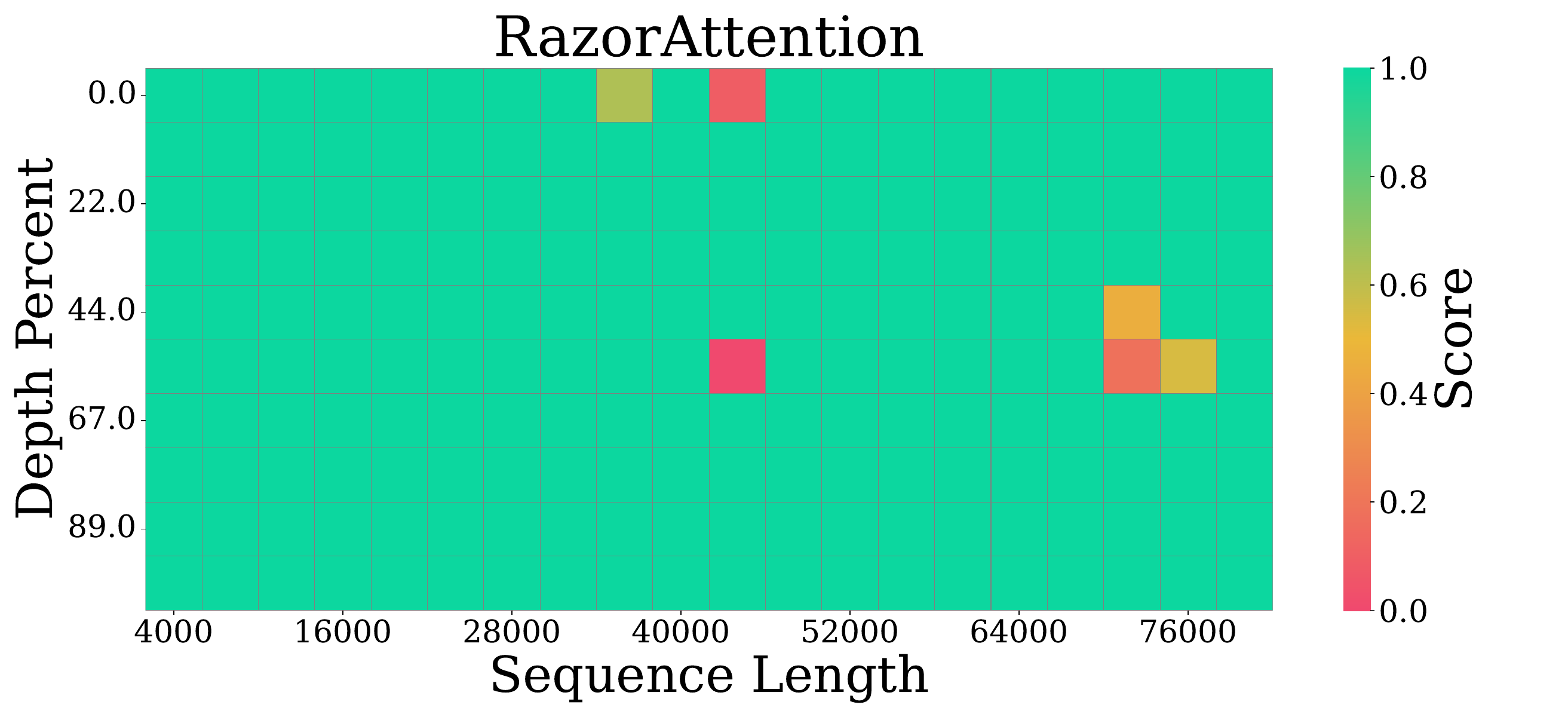}
\includegraphics[width=0.32\textwidth]{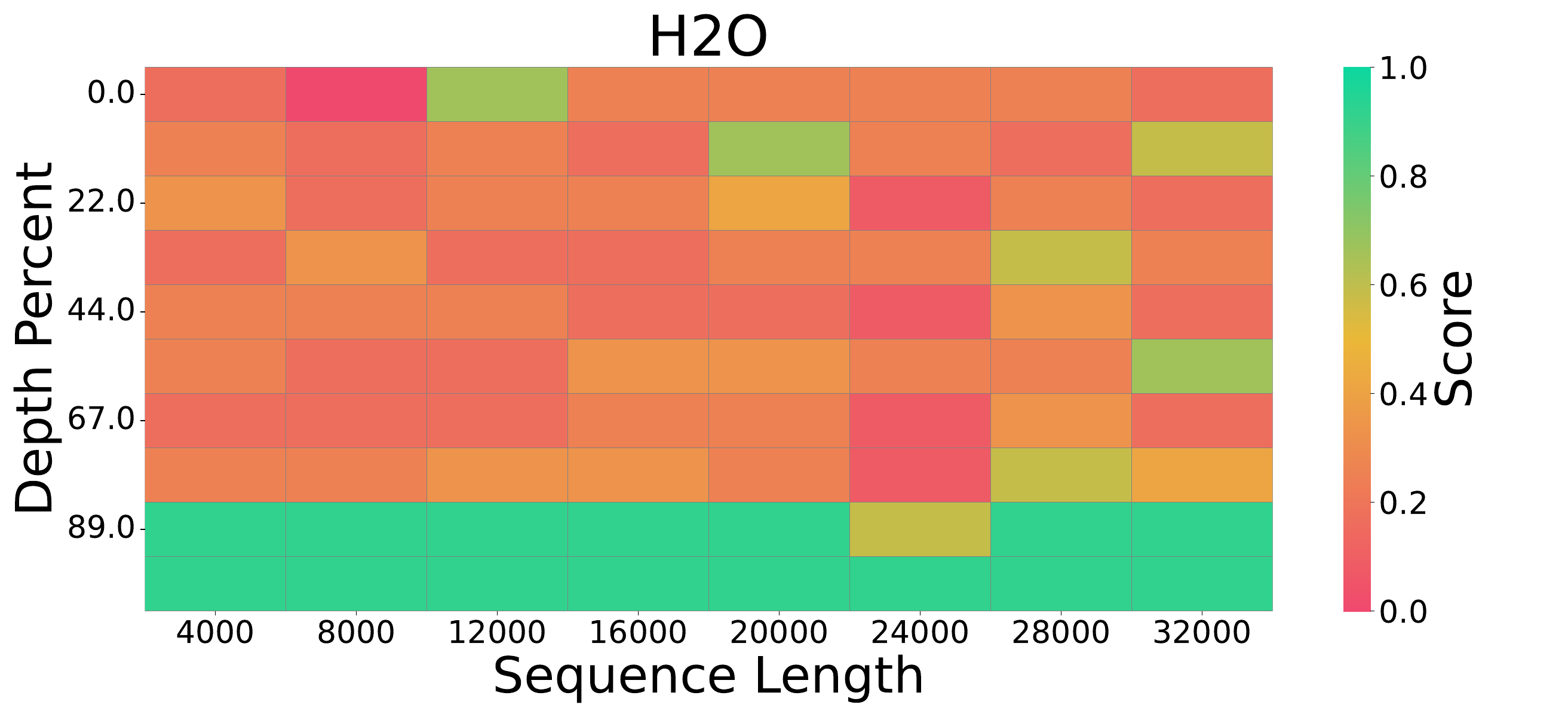}
\caption{Performance comparison of \alg and other compression algorithms on Llama2-7b-80K, Needle In A Haystack. Notice that H2O is incompatible with FlashAttention so we get OOM errors when tested on longer sequences, and its performance has already become unusable in this case.}
\label{fig:llama2_needle}
\end{figure*}

\section{Experiments}
A variety of recent-released LLMs are selected to validate our proposals, including Qwen \citep{bai2023qwen}, Llama2~\citep{llama2}, Llama3 \citep{llama3modelcard} and Baichuan \citep{baichuan2023baichuan2}. The selected models are evaluated on Longbench~\citep{bai2023longbench} and Needle In A Haystack~\citep{gkamradt2023needle} to demonstrate their capabilities in long-context circumstances. The experiments are conducted on NVIDIA GeForce RTX 4090 (24GB). We will first validate the effectiveness of our proposal on various tasks, followed by the ablation study of each component in our algorithm design. Unless explicitly stated, we use \alg with the hyper-parameters as in Table \ref{tab:hyperparameters}. We use H2O~\citep{zhang2023ho} and StreamingLLM~\citep{xiao2024efficient} for comparison. Notice that we do not include SnapKV~\citep{li2024snapkv} as the baseline because it assumes that the query is known before compression, which does not hold in general cases or in a multi-round conversation where the user might query different information from the context (as discussed in Section~\ref{sec:intro}).

\subsection{LongBench Evaluation}
In Table~\ref{tab:longbench} we present the results of different algorithms on LongBench~\citep{bai2023longbench}, which provides a comprehensive assessment to evaluate long-context related abilities of LLMs. We use Qwen1.5-7B and Qwen1.5-72B for testing since  they are RoPE models with a context length of 32K. We also include Llama3-8B to validate the performance of \alg on GQA models. We choose Baichuan2-13B to demonstrate the effectiveness of \alg on ALiBi models. It can be seen that \alg achieved a superior performance across all models compared to StreamingLLM and H2O. The compelling outcomes indicate that  \alg can achieve comparable performance as the uncompressed baseline, even under 3X compression ratio.

Moreover, we test Llama3-8B-Instruct as a GQA instance where every 4 attention heads share a single set of KV cache. Hence, we consider the attention heads in a group as all retrieval if one or more heads satisfy inductive or echoing property. The results in Table~\ref{tab:longbench} clearly prove that \alg still work for GQA models.

\begin{figure}[H]
\centering  
\includegraphics[width=0.49\textwidth]{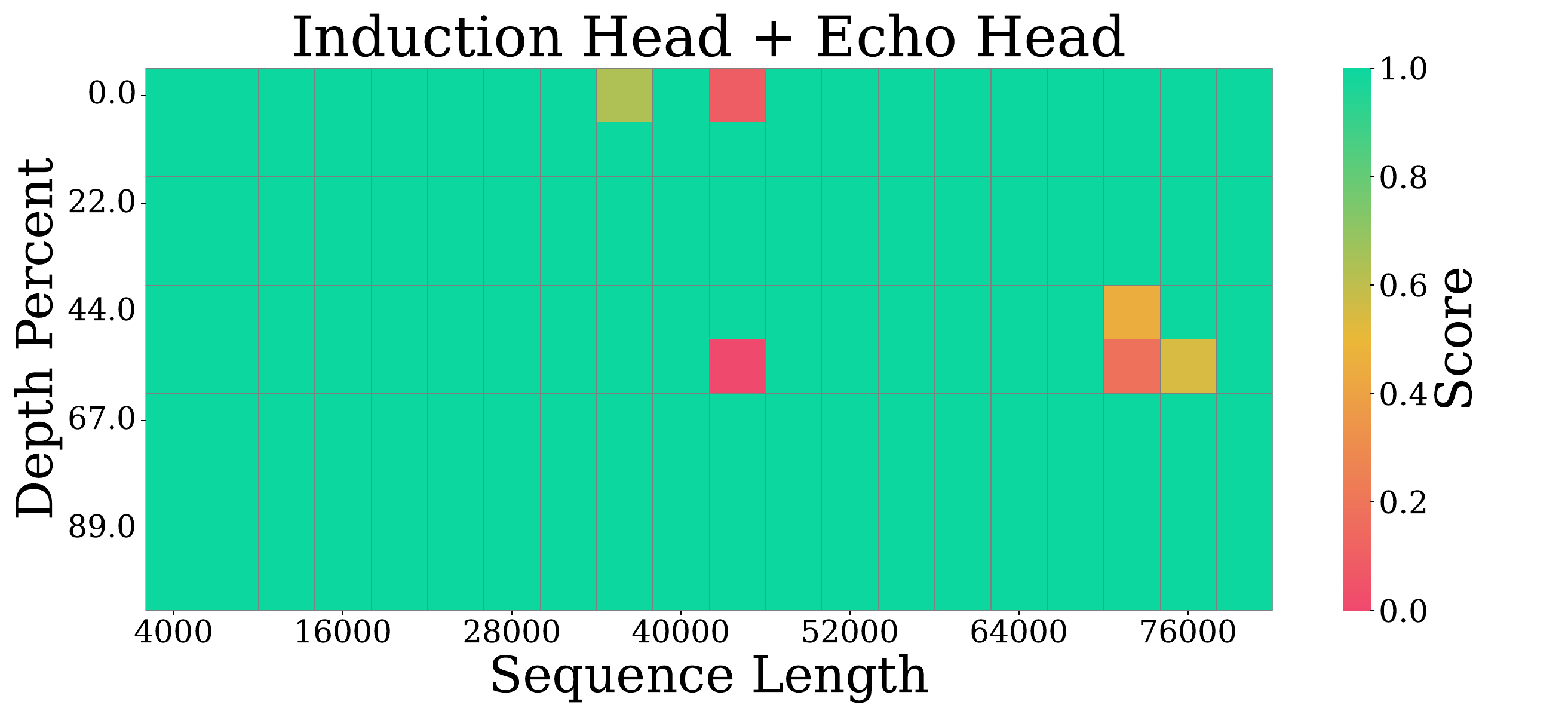}
\includegraphics[width=0.49\textwidth]{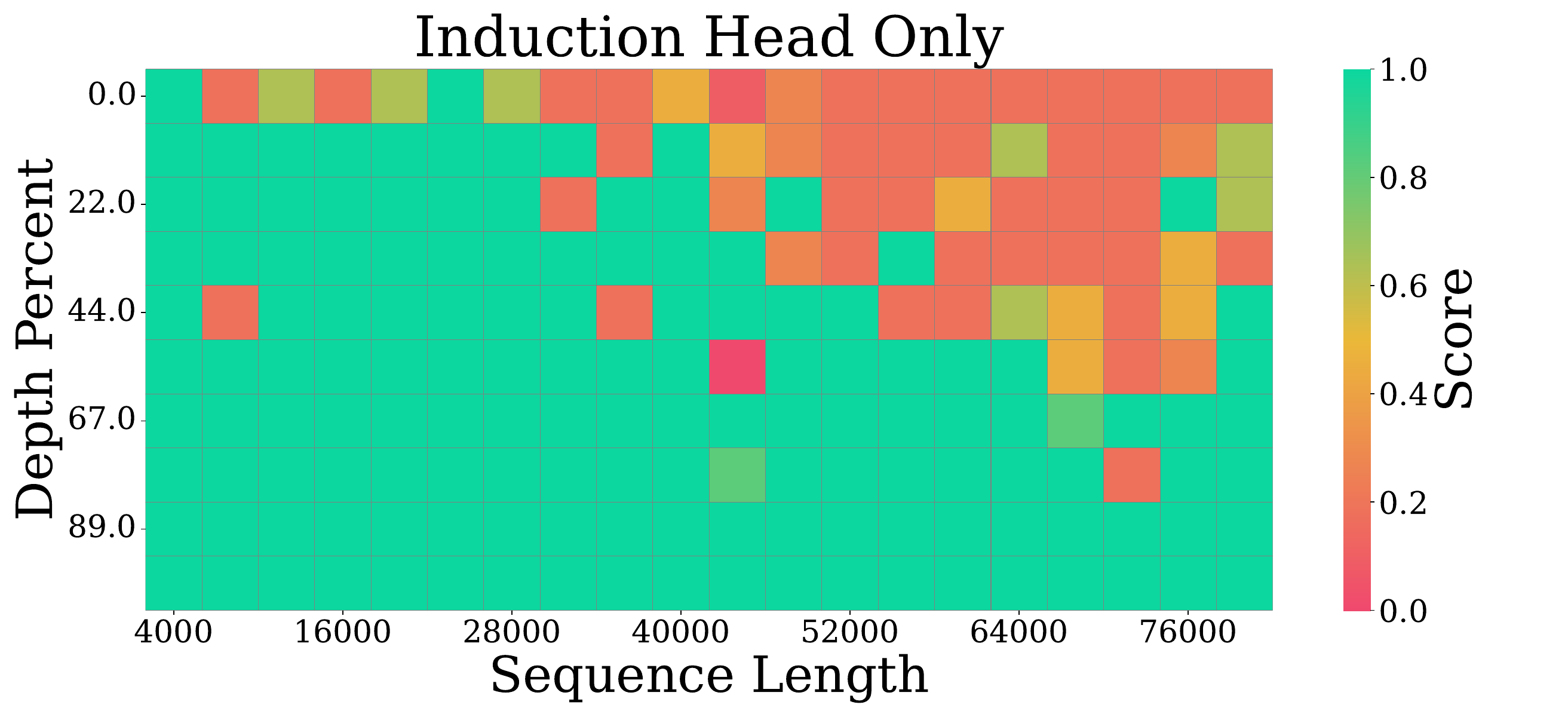}
\caption{Adding $1\%$ of the echo heads can significantly enhances the retrieving performance of \alg on Llama2-7B-80k. }
\label{fig:what_if_no_echo}
\end{figure}

\subsection{Needle In A Haystack Evaluation}
In Figure~\ref{fig:llama2_needle} we present the results on Needle In A Haystack. We use Llama2-7B-80K from ~\cite{fu2024data} since the context length of this model is 80K. Unlike H2O, whose performance is severely degraded under long inputs, \alg can still accurately recall the queried information. This is a strong evidence proving that \alg can retain all the semantic information within the original context, while importance-based methods inevitably discard information that might be useful in future queries.

\subsection{Ablation Studies}\label{sec:ablation}
Below we present the ablation results of \blg, and prove that the algorithm design and configuration are optimally chosen to achieve a higher compression ration with acceptable performance degradation.

\subsubsection{Importance of Echo Heads}
Although we only include $1\%$ echo heads in \blg, we notice that this group of heads is quite essential in retrieving information under long context as shown in Figure~\ref{fig:what_if_no_echo}. One possible explanation is that the induction heads depend on the existence of echo heads as discussed in ~\cite{olsson2022incontext}.

\begin{table}
\centering
\begin{tabular}{|c|c|}
\hline
Protection scheme & Score \\\hline
1\% Echo + 5\% Induction Head & 69.54\%\\
1\% Echo + 8\% Induction Head & 78.40\%\\
1\% Echo + 11\% Induction Head & 84.55\%\\
1\% Echo + 14\% Induction Head & 86.59\%\\
Baseline & 87.05\%
\\\hline
\end{tabular}
\caption{\label{tab:ablation_num_induction}Qwen1.5-7B-Chat using \alg with different numbers of heads protected, tested on Needle in A Haystack.}
\end{table}

\subsubsection{Number of Induction Heads}
To determine the optimal number of induction heads to use in \blg, in Table~\ref{tab:ablation_num_induction} we present the accuracy of \alg under various numbers of induction heads. The results show that the accuracy improves continuously with an increasing number of induction heads. We decide to include $14\%$ of the induction heads in order to achieve an optimal balance between the compression ratio and model performance.

\begin{figure}[H]
\centering  
\includegraphics[width=0.49\textwidth]{llama-2-7b-80k-compress.pdf}
\includegraphics[width=0.49\textwidth]{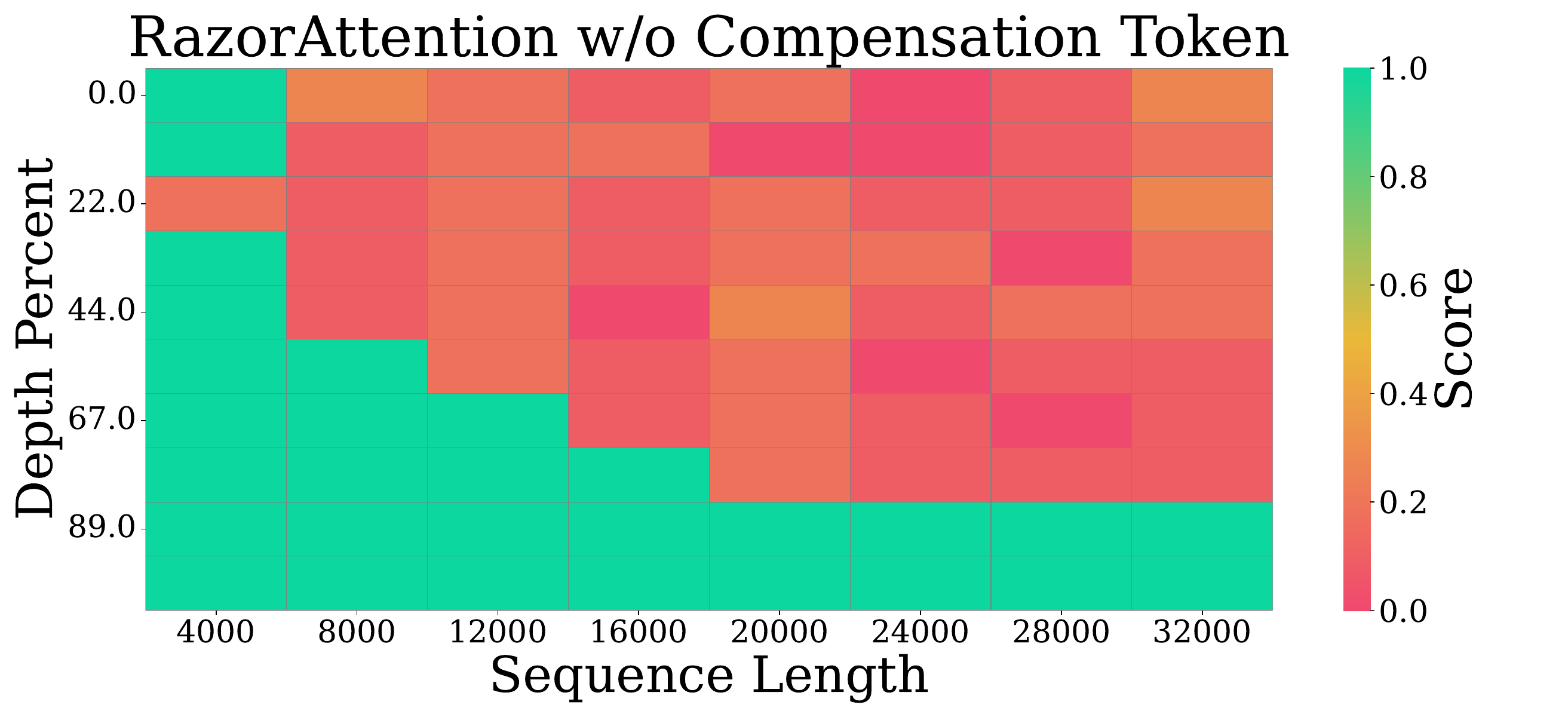}
\caption{The compensation token is critical for recovering the information loss introduced by the truncated KV cache.}
\label{fig:ablation_compenation_token}
\end{figure}

\subsubsection{Importance of the Compensation Token}\label{sec:ablation_compensation_token}
In Figure~\ref{fig:ablation_compenation_token}, it is clearly demonstrated that compensation tokens are critical for the performance of \blg. The compensation tokens successfully compressed most of the information from the dropped tokens,thereby maintaining high accuracy even with significant KV cache reduction.

\section{Conclusion}
In this paper, we propose \blg, a novel KV cache compression algorithm, which successfully achieves a 3X compression ratio for models use RoPE or ALiBi embeddings. Unlike previous importance-based token-dropping methods which inevitably discard semantic information, \alg preserves all semantic information within retrieval heads. We demonstrate that remote tokens can be effectively compressed into compensation tokens within non-retrieval heads. Furthermore, our head-wise pruning criterion is fully compatible with FlashAttention, making \alg a plug-and-play compression method that accelerates the inference of LLMs under extended context. Our experiments demonstrate that \alg can achieve comparable performance with the original model and surpasses previous methods in both accuracy and efficiency.

\section{Limitation}
However, there are still certain limitations of our work. The first question is why attention heads in LLMs behave so differently and how retrieval heads operate under lengthy inputs. The second challenge lies in achieving a higher compression ratio. Although we have successfully reduced the KV cache by $70\%$, we believe this number can be further improved. Moreover, although we have tested our algorithm on several models, the optimal configuration on other models might be different, meaning that we might need more or less retrieval heads under different cases. These topics are quite important and we will keep investigating them in the future work.



\begin{thebibliography}{10}

\bibitem{pmlr-v202-sheng23a}
Ying Sheng, Lianmin Zheng, Binhang Yuan, Zhuohan Li, Max Ryabinin, Beidi Chen,
  Percy Liang, Christopher Re, Ion Stoica, and Ce~Zhang.
\newblock {F}lex{G}en: High-throughput generative inference of large language
  models with a single {GPU}.
\newblock In Andreas Krause, Emma Brunskill, Kyunghyun Cho, Barbara Engelhardt,
  Sivan Sabato, and Jonathan Scarlett, editors, {\em Proceedings of the 40th
  International Conference on Machine Learning}, volume 202 of {\em Proceedings
  of Machine Learning Research}, pages 31094--31116. PMLR, 23--29 Jul 2023.

\bibitem{zhao2024atom}
Yilong Zhao, Chien-Yu Lin, Kan Zhu, Zihao Ye, Lequn Chen, Size Zheng, Luis
  Ceze, Arvind Krishnamurthy, Tianqi Chen, and Baris Kasikci.
\newblock Atom: Low-bit quantization for efficient and accurate llm serving,
  2024.

\bibitem{lin2024qserve}
Yujun Lin, Haotian Tang, Shang Yang, Zhekai Zhang, Guangxuan Xiao, Chuang Gan,
  and Song Han.
\newblock Qserve: W4a8kv4 quantization and system co-design for efficient llm
  serving, 2024.

\bibitem{zhang2023ho}
Zhenyu Zhang, Ying Sheng, Tianyi Zhou, Tianlong Chen, Lianmin Zheng, Ruisi Cai,
  Zhao Song, Yuandong Tian, Christopher Re, Clark Barrett, Zhangyang Wang, and
  Beidi Chen.
\newblock H2o: Heavy-hitter oracle for efficient generative inference of large
  language models.
\newblock In {\em Thirty-seventh Conference on Neural Information Processing
  Systems}, 2023.

\bibitem{xiao2024efficient}
Guangxuan Xiao, Yuandong Tian, Beidi Chen, Song Han, and Mike Lewis.
\newblock Efficient streaming language models with attention sinks, 2024.

\bibitem{mistral}
Albert~Q. Jiang, Alexandre Sablayrolles, Arthur Mensch, Chris Bamford,
  Devendra~Singh Chaplot, Diego de~las Casas, Florian Bressand, Gianna Lengyel,
  Guillaume Lample, Lucile Saulnier, Lélio~Renard Lavaud, Marie-Anne Lachaux,
  Pierre Stock, Teven~Le Scao, Thibaut Lavril, Thomas Wang, Timothée Lacroix,
  and William~El Sayed.
\newblock Mistral 7b, 2023.

\bibitem{sparseattn}
Rewon Child, Scott Gray, Alec Radford, and Ilya Sutskever.
\newblock Generating long sequences with sparse transformers, 2019.

\bibitem{fu2024data}
Yao Fu, Rameswar Panda, Xinyao Niu, Xiang Yue, Hannaneh Hajishirzi, Yoon Kim,
  and Hao Peng.
\newblock Data engineering for scaling language models to 128k context, 2024.

\bibitem{gkamradt2023needle}
gkamradt.
\newblock {Needle In A Haystack - Pressure Testing LLMs}, 2023.

\bibitem{bai2023longbench}
Yushi Bai, Xin Lv, Jiajie Zhang, Hongchang Lyu, Jiankai Tang, Zhidian Huang,
  Zhengxiao Du, Xiao Liu, Aohan Zeng, Lei Hou, Yuxiao Dong, Jie Tang, and
  Juanzi Li.
\newblock Longbench: A bilingual, multitask benchmark for long context
  understanding, 2023.

\bibitem{liu2023scissorhands}
Zichang Liu, Aditya Desai, Fangshuo Liao, Weitao Wang, Victor Xie, Zhaozhuo Xu,
  Anastasios Kyrillidis, and Anshumali Shrivastava.
\newblock Scissorhands: Exploiting the persistence of importance hypothesis for
  llm kv cache compression at test time, 2023.

\bibitem{li2024snapkv}
Yuhong Li, Yingbing Huang, Bowen Yang, Bharat Venkitesh, Acyr Locatelli,
  Hanchen Ye, Tianle Cai, Patrick Lewis, and Deming Chen.
\newblock Snapkv: Llm knows what you are looking for before generation, 2024.

\bibitem{bai2023qwen}
Jinze Bai, Shuai Bai, Yunfei Chu, Zeyu Cui, Kai Dang, Xiaodong Deng, Yang Fan,
  Wenbin Ge, Yu~Han, Fei Huang, Binyuan Hui, Luo Ji, Mei Li, Junyang Lin, Runji
  Lin, Dayiheng Liu, Gao Liu, Chengqiang Lu, Keming Lu, Jianxin Ma, Rui Men,
  Xingzhang Ren, Xuancheng Ren, Chuanqi Tan, Sinan Tan, Jianhong Tu, Peng Wang,
  Shijie Wang, Wei Wang, Shengguang Wu, Benfeng Xu, Jin Xu, An~Yang, Hao Yang,
  Jian Yang, Shusheng Yang, Yang Yao, Bowen Yu, Hongyi Yuan, Zheng Yuan,
  Jianwei Zhang, Xingxuan Zhang, Yichang Zhang, Zhenru Zhang, Chang Zhou,
  Jingren Zhou, Xiaohuan Zhou, and Tianhang Zhu.
\newblock Qwen technical report, 2023.

\bibitem{llama2}
Hugo Touvron, Louis Martin, Kevin Stone, Peter Albert, Amjad Almahairi, Yasmine
  Babaei, Nikolay Bashlykov, Soumya Batra, Prajjwal Bhargava, Shruti Bhosale,
  Dan Bikel, Lukas Blecher, Cristian~Canton Ferrer, Moya Chen, Guillem
  Cucurull, David Esiobu, Jude Fernandes, Jeremy Fu, Wenyin Fu, Brian Fuller,
  Cynthia Gao, Vedanuj Goswami, Naman Goyal, Anthony Hartshorn, Saghar
  Hosseini, Rui Hou, Hakan Inan, Marcin Kardas, Viktor Kerkez, Madian Khabsa,
  Isabel Kloumann, Artem Korenev, Punit~Singh Koura, Marie-Anne Lachaux,
  Thibaut Lavril, Jenya Lee, Diana Liskovich, Yinghai Lu, Yuning Mao, Xavier
  Martinet, Todor Mihaylov, Pushkar Mishra, Igor Molybog, Yixin Nie, Andrew
  Poulton, Jeremy Reizenstein, Rashi Rungta, Kalyan Saladi, Alan Schelten, Ruan
  Silva, Eric~Michael Smith, Ranjan Subramanian, Xiaoqing~Ellen Tan, Binh Tang,
  Ross Taylor, Adina Williams, Jian~Xiang Kuan, Puxin Xu, Zheng Yan, Iliyan
  Zarov, Yuchen Zhang, Angela Fan, Melanie Kambadur, Sharan Narang, Aurelien
  Rodriguez, Robert Stojnic, Sergey Edunov, and Thomas Scialom.
\newblock Llama 2: Open foundation and fine-tuned chat models, 2023.

\bibitem{llama3modelcard}
AI@Meta.
\newblock Llama 3 model card.
\newblock 2024.

\bibitem{baichuan2023baichuan2}
Baichuan.
\newblock Baichuan 2: Open large-scale language models.
\newblock {\em arXiv preprint arXiv:2309.10305}, 2023.

\bibitem{mamba}
Albert Gu and Tri Dao.
\newblock Mamba: Linear-time sequence modeling with selective state spaces,
  2024.

\bibitem{mamba2}
Tri Dao and Albert Gu.
\newblock Transformers are ssms: Generalized models and efficient algorithms
  through structured state space duality, 2024.

\bibitem{infinitransformers}
Tsendsuren Munkhdalai, Manaal Faruqui, and Siddharth Gopal.
\newblock Leave no context behind: Efficient infinite context transformers with
  infini-attention, 2024.

\bibitem{rwkv}
Bo~Peng, Eric Alcaide, Quentin Anthony, Alon Albalak, Samuel Arcadinho, Stella
  Biderman, Huanqi Cao, Xin Cheng, Michael Chung, Matteo Grella, Kranthi~Kiran
  GV, Xuzheng He, Haowen Hou, Jiaju Lin, Przemyslaw Kazienko, Jan Kocon,
  Jiaming Kong, Bartlomiej Koptyra, Hayden Lau, Krishna Sri~Ipsit Mantri,
  Ferdinand Mom, Atsushi Saito, Guangyu Song, Xiangru Tang, Bolun Wang,
  Johan~S. Wind, Stanislaw Wozniak, Ruichong Zhang, Zhenyuan Zhang, Qihang
  Zhao, Peng Zhou, Qinghua Zhou, Jian Zhu, and Rui-Jie Zhu.
\newblock Rwkv: Reinventing rnns for the transformer era, 2023.

\bibitem{griffin}
Soham De, Samuel~L. Smith, Anushan Fernando, Aleksandar Botev, George
  Cristian-Muraru, Albert Gu, Ruba Haroun, Leonard Berrada, Yutian Chen,
  Srivatsan Srinivasan, Guillaume Desjardins, Arnaud Doucet, David Budden,
  Yee~Whye Teh, Razvan Pascanu, Nando~De Freitas, and Caglar Gulcehre.
\newblock Griffin: Mixing gated linear recurrences with local attention for
  efficient language models, 2024.

\bibitem{pmlr-v202-xiao23c}
Guangxuan Xiao, Ji~Lin, Mickael Seznec, Hao Wu, Julien Demouth, and Song Han.
\newblock {S}mooth{Q}uant: Accurate and efficient post-training quantization
  for large language models.
\newblock In Andreas Krause, Emma Brunskill, Kyunghyun Cho, Barbara Engelhardt,
  Sivan Sabato, and Jonathan Scarlett, editors, {\em Proceedings of the 40th
  International Conference on Machine Learning}, volume 202 of {\em Proceedings
  of Machine Learning Research}, pages 38087--38099. PMLR, 23--29 Jul 2023.

\bibitem{NEURIPS2022_6f6db140}
Xiuying Wei, Yunchen Zhang, Xiangguo Zhang, Ruihao Gong, Shanghang Zhang,
  Qi~Zhang, Fengwei Yu, and Xianglong Liu.
\newblock Outlier suppression: Pushing the limit of low-bit transformer
  language models.
\newblock In S.~Koyejo, S.~Mohamed, A.~Agarwal, D.~Belgrave, K.~Cho, and A.~Oh,
  editors, {\em Advances in Neural Information Processing Systems}, volume~35,
  pages 17402--17414. Curran Associates, Inc., 2022.

\bibitem{wei2023outlier}
Xiuying Wei, Yunchen Zhang, Yuhang Li, Xiangguo Zhang, Ruihao Gong, Jinyang
  Guo, and Xianglong Liu.
\newblock Outlier suppression+: Accurate quantization of large language models
  by equivalent and optimal shifting and scaling, 2023.

\bibitem{hooper2024kvquant}
Coleman Hooper, Sehoon Kim, Hiva Mohammadzadeh, Michael~W. Mahoney,
  Yakun~Sophia Shao, Kurt Keutzer, and Amir Gholami.
\newblock Kvquant: Towards 10 million context length llm inference with kv
  cache quantization, 2024.

\bibitem{liu2024kivi}
{Zirui Liu}, {Jiayi Yuan}, {Hongye Jin}, {Shaochen Zhong}, {Zhaozhuo Xu},
  Vladimir Braverman, {Beidi Chen}, and Xia Hu.
\newblock Kivi : Plug-and-play 2bit kv cache quantization with streaming
  asymmetric quantization.
\newblock 2023.

\bibitem{liu2023llm}
Zechun Liu, Barlas Oguz, Changsheng Zhao, Ernie Chang, Pierre Stock, Yashar
  Mehdad, Yangyang Shi, Raghuraman Krishnamoorthi, and Vikas Chandra.
\newblock Llm-qat: Data-free quantization aware training for large language
  models.
\newblock {\em arXiv preprint arXiv:2305.17888}, 2023.

\bibitem{cai2024pyramidkv}
Zefan Cai., Yichi Zhang, Bofei Gao, Tianyu Liu, Keming Lu, Wayne Xiong, Yue
  Dong, Baobao Chang, Junjie Hu, and Wen Xiao.
\newblock Pyramidkv: Dynamic kv cache compression based on pyramidal
  information funneling, 2024.

\bibitem{yang2024pyramidinfer}
Dongjie Yang, XiaoDong Han, Yan Gao, Yao Hu, Shilin Zhang, and Hai Zhao.
\newblock Pyramidinfer: Pyramid kv cache compression for high-throughput llm
  inference, 2024.

\bibitem{ge2024model}
Suyu Ge, Yunan Zhang, Liyuan Liu, Minjia Zhang, Jiawei Han, and Jianfeng Gao.
\newblock Model tells you what to discard: Adaptive kv cache compression for
  llms, 2024.

\bibitem{zandieh2024subgen}
Amir Zandieh, Insu Han, Vahab Mirrokni, and Amin Karbasi.
\newblock Subgen: Token generation in sublinear time and memory, 2024.

\bibitem{dai2024sequence}
Jincheng Dai, Zhuowei Huang, Haiyun Jiang, Chen Chen, Deng Cai, Wei Bi, and
  Shuming Shi.
\newblock Sequence can secretly tell you what to discard, 2024.

\bibitem{shazeer2019fast}
Noam Shazeer.
\newblock Fast transformer decoding: One write-head is all you need, 2019.

\bibitem{ainslie2023gqa}
Joshua Ainslie, James Lee-Thorp, Michiel de~Jong, Yury Zemlyanskiy, Federico
  Lebrón, and Sumit Sanghai.
\newblock Gqa: Training generalized multi-query transformer models from
  multi-head checkpoints, 2023.

\bibitem{deepseekai2024deepseekv2}
DeepSeek-AI, Aixin Liu, Bei Feng, Bin Wang, Bingxuan Wang, Bo~Liu, Chenggang
  Zhao, Chengqi Dengr, Chong Ruan, Damai Dai, Daya Guo, Dejian Yang, Deli Chen,
  Dongjie Ji, Erhang Li, Fangyun Lin, Fuli Luo, Guangbo Hao, Guanting Chen,
  Guowei Li, H.~Zhang, Hanwei Xu, Hao Yang, Haowei Zhang, Honghui Ding, Huajian
  Xin, Huazuo Gao, Hui Li, Hui Qu, J.~L. Cai, Jian Liang, Jianzhong Guo, Jiaqi
  Ni, Jiashi Li, Jin Chen, Jingyang Yuan, Junjie Qiu, Junxiao Song, Kai Dong,
  Kaige Gao, Kang Guan, Lean Wang, Lecong Zhang, Lei Xu, Leyi Xia, Liang Zhao,
  Liyue Zhang, Meng Li, Miaojun Wang, Mingchuan Zhang, Minghua Zhang, Minghui
  Tang, Mingming Li, Ning Tian, Panpan Huang, Peiyi Wang, Peng Zhang, Qihao
  Zhu, Qinyu Chen, Qiushi Du, R.~J. Chen, R.~L. Jin, Ruiqi Ge, Ruizhe Pan,
  Runxin Xu, Ruyi Chen, S.~S. Li, Shanghao Lu, Shangyan Zhou, Shanhuang Chen,
  Shaoqing Wu, Shengfeng Ye, Shirong Ma, Shiyu Wang, Shuang Zhou, Shuiping Yu,
  Shunfeng Zhou, Size Zheng, T.~Wang, Tian Pei, Tian Yuan, Tianyu Sun, W.~L.
  Xiao, Wangding Zeng, Wei An, Wen Liu, Wenfeng Liang, Wenjun Gao, Wentao
  Zhang, X.~Q. Li, Xiangyue Jin, Xianzu Wang, Xiao Bi, Xiaodong Liu, Xiaohan
  Wang, Xiaojin Shen, Xiaokang Chen, Xiaosha Chen, Xiaotao Nie, Xiaowen Sun,
  Xiaoxiang Wang, Xin Liu, Xin Xie, Xingkai Yu, Xinnan Song, Xinyi Zhou, Xinyu
  Yang, Xuan Lu, Xuecheng Su, Y.~Wu, Y.~K. Li, Y.~X. Wei, Y.~X. Zhu, Yanhong
  Xu, Yanping Huang, Yao Li, Yao Zhao, Yaofeng Sun, Yaohui Li, Yaohui Wang,
  Yi~Zheng, Yichao Zhang, Yiliang Xiong, Yilong Zhao, Ying He, Ying Tang, Yishi
  Piao, Yixin Dong, Yixuan Tan, Yiyuan Liu, Yongji Wang, Yongqiang Guo, Yuchen
  Zhu, Yuduan Wang, Yuheng Zou, Yukun Zha, Yunxian Ma, Yuting Yan, Yuxiang You,
  Yuxuan Liu, Z.~Z. Ren, Zehui Ren, Zhangli Sha, Zhe Fu, Zhen Huang, Zhen
  Zhang, Zhenda Xie, Zhewen Hao, Zhihong Shao, Zhiniu Wen, Zhipeng Xu, Zhongyu
  Zhang, Zhuoshu Li, Zihan Wang, Zihui Gu, Zilin Li, and Ziwei Xie.
\newblock Deepseek-v2: A strong, economical, and efficient mixture-of-experts
  language model, 2024.

\bibitem{olsson2022incontext}
Catherine Olsson, Nelson Elhage, Neel Nanda, Nicholas Joseph, Nova DasSarma,
  Tom Henighan, Ben Mann, Amanda Askell, Yuntao Bai, Anna Chen, Tom Conerly,
  Dawn Drain, Deep Ganguli, Zac Hatfield-Dodds, Danny Hernandez, Scott
  Johnston, Andy Jones, Jackson Kernion, Liane Lovitt, Kamal Ndousse, Dario
  Amodei, Tom Brown, Jack Clark, Jared Kaplan, Sam McCandlish, and Chris Olah.
\newblock In-context learning and induction heads, 2022.

\bibitem{wu2024retrieval}
Wenhao Wu, Yizhong Wang, Guangxuan Xiao, Hao Peng, and Yao Fu.
\newblock Retrieval head mechanistically explains long-context factuality,
  2024.

\bibitem{alibi}
Ofir Press, Noah~A. Smith, and Mike Lewis.
\newblock Train short, test long: Attention with linear biases enables input
  length extrapolation, 2022.

\bibitem{rope}
Jianlin Su, Yu~Lu, Shengfeng Pan, Ahmed Murtadha, Bo~Wen, and Yunfeng Liu.
\newblock Roformer: Enhanced transformer with rotary position embedding, 2023.

\bibitem{rmsnorm}
Biao Zhang and Rico Sennrich.
\newblock Root mean square layer normalization, 2019.

\end{thebibliography}

\newpage
\appendix
\section{Appendix: Proof of Theorem~\ref{theo:ALiBi_theo}}
\label{sec:proof}
Below we first give an upper bound for the product of the queries and keys, and then show that the attention weight would decay to zero when  the positional bias is significantly larger than that upper bound.
\begin{proof}
Since we have $\bm q = W_{Q_h}\bm x $ and $\bm k = W_{K_h} \bm x$ where $\bm x $ is the input of the Attention block, this leads to 
\begin{align*}
\bm q \bm k^{\intercal} = \bm x W_{Q_h}W_{K_h}\bm x^{\intercal} \leq \|W_{Q_h}W_{K_h}\|_2\|\bm x\|^2.\numberthis\label{app:eq_qk_upper_bound}
\end{align*}
Since $\bm x$ is attained after LayerNorm, which means
\begin{align*}
&\bm x = \bm \gamma \odot \frac{\bm \hat{x} - \mu}{\sigma} + \bm b,\\
&\mu = \frac{1}{d}\sum_{i=1}^d \hat{x}_i,\quad \sigma = \frac{1}{d}\sum_{i=1}^d(\hat{x}_i - \mu)^2.
\end{align*}
Here $\bm \hat{x}$ is the input of LayerNorm, $d$ is its dimension and $\hat{x}_i$ is the $i$-th dimension of $\bm \hat{x}$. The equation above leads to
\begin{align*}
\|\bm x\|^2 = &\left\|\bm \gamma \odot \frac{\bm \hat{x} - \mu}{\sigma} + \bm b\right\|^2\\
\leq & 2\left\|\bm \gamma \odot \frac{\bm \hat{x} - \mu}{\sigma}\right\|^2 + 2\|\bm b\|^2\\
\leq & 2\|\bm \gamma\|^2 + 2\|\bm b\|^2.\numberthis\label{app:eq_x_upper_bound}
\end{align*}
Combining \eqref{app:eq_qk_upper_bound} and \eqref{app:eq_x_upper_bound} we get
\begin{align*}
\bm q \bm k^{\intercal} \leq \|W_{Q_h}W_{K_h}\|_2\left( 2\|\bm \gamma\|^2 + 2\|\bm b\|^2 \right)\numberthis\label{app:upper_qk_2}
\end{align*}
In order to give an upper bound for the attention weight, we have
\begin{align*}
\text{Attn}_{m\to n}\left(\bm{q};\bm{k}  \right) =& \frac{\exp{\left(S_{m\to n}\left(\bm{q};\bm{k}  \right) \right)}}{\sum_{n=0}^{m}\exp{\left(S_{m\to n}\left(\bm{q};\bm{k}  \right) \right)}}\\
= & \frac{\exp{\left(\bm{q}\bm{k}^{\intercal} - l_h(m-n) \right) }}{\sum_{n=0}^{m}\exp{\left(S_{m\to n}\left(\bm{q};\bm{k}  \right) \right)}}\\
\leq & \frac{\exp{\left(\bm{q}\bm{k}^{\intercal} - l_h(m-n) \right) }}{\exp{\left(S_{n\to n}\left(\bm{q};\bm{k}  \right) \right)}}\\
\leq & \frac{\exp{\left(\bm{q}\bm{k}^{\intercal} - l_h(m-n) \right) }}{\exp{\left( \bm q\bm q^{\intercal}\right)}}\\
\leq & \exp{\left(\bm{q}\bm{k}^{\intercal} - l_h(m-n) \right)}\\
= & \frac{\exp{\left(\bm{q}\bm{k}^{\intercal}\right)}}{\exp{\left(l_h(m-n) \right)}}.
\end{align*}
Therefore to ensure $\text{Attn}_{m\to n}\left(\bm{q};\bm{k}  \right)\leq \epsilon$, which is equivalent as $\log\left(\text{Attn}_{m\to n}\left(\bm{q};\bm{k}  \right)\right)\leq \log(\epsilon)$, we need
\begin{align*}
\log\left(\text{Attn}_{m\to n}\left(\bm{q};\bm{k}  \right)\right) \leq & \bm{q}\bm{k}^{\intercal} - l_h(m-n) \leq \log(\epsilon)
\end{align*}
Taking \eqref{app:upper_qk_2} into the equation above, we get
\begin{align*}
\|W_{Q_h}W_{K_h}\|_2\left( 2\|\bm \gamma\|^2 + 2\|\bm b\|^2 \right) - l_h(m-n) \leq \log(\epsilon),
\end{align*}
which gives us
\begin{align*}
m - n \geq \frac{2\|W_{Q_h}W_{K_h}\|_2\left( \|\bm \gamma\|^2 + \|\bm b\|^2 \right) - \log(\epsilon)}{l_h}.
\end{align*}
In this case, we have 
\begin{align*}
\text{Attn}_{m\to n}\left(\bm{q};\bm{k}  \right) \leq \epsilon.
\end{align*}
\end{proof}

\end{document}